\documentclass[runningheads]{llncs}

\usepackage[T1]{fontenc}

\usepackage{hyperref}
\usepackage{url}

\usepackage{booktabs}

\usepackage{mathtools}
\usepackage{amssymb}

\newcommand{\bftab}{\fontseries{b}\selectfont}
\newcommand{\e}[1]{\emph{#1}}

\newcommand{\ra}{\rightarrow}

\DeclarePairedDelimiter\size{\lvert}{\rvert}

\DeclarePairedDelimiter\brackets{(}{)}

\usepackage{wasysym}
\usepackage{scalerel}

\newcommand\bvert[1][\relax]{\makebox[\widthof{$#1\vert$}]{\scalerel*[1.2pt]{\brokenvert}{\raisebox{-.2pt}{\vstretch{1.04}{#1\vert}}}}}
\newcommand\lbvert[1][\relax]{\mathopen{\bvert[#1]}}
\newcommand\rbvert[1][\relax]{\mathclose{\bvert[#1]}}

\makeatletter
\newcommand\lrbvert{%
  \@ifstar{\lrbvertscaled}{\lrbvertfixed}%
}
\newcommand\lrbvertscaled[2][\relax]{%
  \mathopen{\ThisStyle{\mathmakebox[\widthof{$\SavedStyle\left\lvert\mathclap{#2}\right.$}]{\scalerel*[1.2pt]{\brokenvert}{\raisebox{-.2pt}{$\SavedStyle\vstretch{1.04}{\left\lvert#2\right\rvert}$}}}}}%
  #2%
  \mathclose{\ThisStyle{\mathmakebox[\widthof{$\SavedStyle\left.\mathclap{#2}\right\rvert$}]{\scalerel*[1.2pt]{\brokenvert}{\raisebox{-.2pt}{$\SavedStyle\vstretch{1.04}{\left\lvert#2\right\rvert}$}}}}}%
}
\newcommand\lrbvertfixed[2][\relax]{%
  \lbvert[#1]#2\rbvert[#1]%
}
\makeatother
\newcommand\sgnpow\lrbvert

\RequirePackage{color}\definecolor{RED}{rgb}{1,0,0}

\begin{document}

\title{Monotonic anomaly detection}

\author{Oliver Urs Lenz\inst{1}\inst{2}\inst{3}%\orcidID{0000-1111-2222-3333}
\and
Matthijs van Leeuwen\inst{2}%\orcidID{0000-1111-2222-3333}
}
\authorrunning{O U Lenz \& M van Leeuwen}
\institute{Department of Statistics, Data Science and Modelling, National Institute for Public Health and the Environment, The Netherlands \and
Leiden Institute of Advanced Computer Science, Leiden University \email{o.u.lenz@liacs.leidenuniv.nl} \and
Department of Applied Mathematics, Computer Science and Statistics,\\Ghent University}

\maketitle

\begin{abstract}%   <- trailing '%' for backward compatibility of .sty file

Semi-supervised anomaly detection is based on the principle that any record that looks different from normal training data is a potential anomaly. However, in some cases we are specifically interested in anomalies that correspond to high attribute values (or low, but not both). For distance-based methods, we propose an asymmetrical distance measure that takes this monotonicity into account by incorporating the ramp function. For the Isolation Forest algorithm, we propose a modified path length algorithm. Through experiments on synthetic and real-life datasets, we show that these proposals increase anomaly detection performance on datasets with monotonic attributes.

\end{abstract}


\section{Introduction}

The defining characteristic of semi-supervised anomaly detection (also known as \e{one-class classification} \cite{tax01oneclass}) is that the training set only contains normal records, but that we still want to create a model that can distinguish between normal records and anomalies. Potential reasons for working in this setting include not wanting to assume that the available examples of anomalies are representative of all anomalies, and wanting to flag all new records that deviate from normal data as potential anomalies. Examples of application domains of anomaly detection include failure detection in industrial settings and screening patients in healthcare.

In the present paper, we are concerned with numerical tabular data, where the training set consists of a representative sample of a certain normal target class. In this context, all semi-supervised anomaly detection algorithms are necessarily based on the principle that an anomaly is any record that is sufficiently different from the normal training data. However, there are circumstances in which we possess additional helpful domain knowledge.

\begin{figure}
\centering
\includegraphics[width=.6\linewidth]{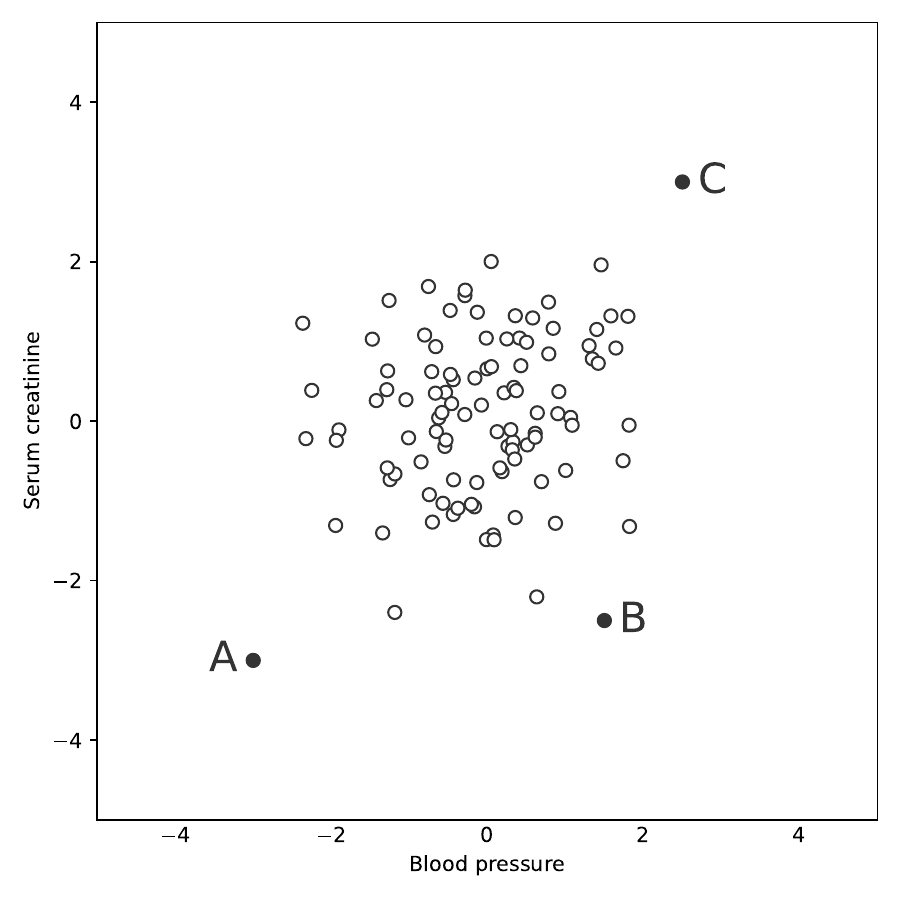}
\caption{Hypothetical example of a medical condition for which high blood pressure and high serum creatinine are known risk factors. Our domain knowledge tells us that only patient C is at-risk, but an ordinary anomaly detector will also flag patients A and B.}
\label{fig_motivating_example}
\end{figure}

Specifically, we consider how best to deal with the knowledge that only relatively high values of an attribute should be indicative of anomality, not relatively low values (or vice-versa). For instance, this applies when certain attributes correspond to higher strain of a machine, or encode known risk factors for patients, and we are interested in detecting machine failure and at-risk patients, rather than underutilisation and exceptionally healthy patients (Fig.~\ref{fig_motivating_example}). We call this setting \e{monotonic} anomaly detection, because our domain knowledge places a monotonicity constraint on certain attributes and the prediction. {Note that this is a soft constraint, since we cannot expect normal records and anomalies to be cleanly separated on every attribute.

When applied to such a monotonic anomaly detection problem, an ordinary anomaly detection algorithm will `detect' a large number of false positives, for it will also flag records representing underutilisation or exceptionally healthy patients, and more generally any records with a sufficiently atypical mixture of relatively high and low attribute values. These false positives create additional work for the practitioner who has to screen potential anomalies, and thereby limit the usefulness of the model. And while it might be possible to use the monotonicity constraints to filter out some clear-cut false positives automatically, coming up with an automated approach to decide whether a flagged record with a mixture of relatively high and low attribute values is a false or a true positive would require reverse-engineering the anomaly detection algorithm.

We propose a more principled approach: modifying existing methods such that relatively low test record values on a monotonic attribute do not lead to an increase in the anomaly score. We will consider six anomaly detectors in this paper. Average Localised Proximity (ALP) \cite{lenz21average}, Support Vector Machine (SVM) \cite{scholkopf99estimating,tax99data}, (weighted) Nearest Neighbour Distance (NND) \cite{knorr97unified,lenz23fuzzy} and Local Outlier Factor (LOF) \cite{breunig00lof} are the four algorithms that performed best in an experimental evaluation across 246 semi-supervised anomaly detection problems \cite{lenz21average}. We also include Centre Distance (CD), i.e. the distance to the centre of the training set, as a simple baseline. These five anomaly detectors are all based on a distance measure in the attribute space, which allows them to extrapolate beyond the normal training data. For these, we propose a modified, asymmetric variant of the commonly used family of Minkowski $p$-distances. Finally, we will also propose a monotonic variant of Isolation Forest (IF) \cite{liu08isolation}, which is a prominent example of an anomaly detector not based on distance.

In Section~\ref{sec_related_work}, we provide a brief overview of some related work. Next, we present our proposal in Section~\ref{sec_dad}. We describe our experimental setup in Section~\ref{sec_experimental_setup} and present the results in Section~\ref{sec_results}, before concluding in Section~\ref{sec_conclusion}.

\section{Related work}
\label{sec_related_work}

There exists a substantial literature on monotonic classification \cite{cano19monotonic}, including monotonic variants of nearest neighbour \cite{duivesteijn08nearest} and decision tree algorithms \cite{kampvande09isotonic,pei18partially,sadawial23monotone}. However, these approaches all rely on a labelled training set, making them fundamentally incompatible with the semi-supervised anomaly detection setting.

To the best of our knowledge, the only previous work discussing something like monotonic anomaly detection, albeit in an unsupervised setting, is the proposal to use the empirical cumulative density function (ECDF) for anomaly detection \cite{li20copod,li22ecod}.\footnote{Somewhat confusingly, the authors have used both \e{copula-based outlier detection} (COPOD) \cite{li20copod} and \e{empirical-cumulative-distribution-based outlier detection} (ECOD) \cite{li22ecod} to refer to their proposal, and included identical implementations under these two different names in their widely-used Python library PyOD \cite{zhao19pyod}.} For every attribute, this approach calculates the two-sided $p$-value of a given test record value, i.e. the probability that the inferred distribution should produce a value at least as extreme as the test record value. The authors note that if anomalies are known to be distributed on one side of the normal data, detection performance is improved if one uses the corresponding one-sided $p$-value. We include ECDF in our experimental evaluation (Section~\ref{sec_results}).

While the proposal on \e{direction-related anomaly detection} \cite{tu24ordinal} appears superficially similar to the present paper, the problem setting in that work is actually quite different from semi-supervised anomaly detection. It assumes that the normal records in the training set have ordinal labels, and uses an ordinal regression algorithm to learn a decision boundary beyond which lie the anomalies. Thus, the directionality in that setting is determined by the response values, and there are no monotonicity constraints on the independent attributes.

Other attempts to incorporate domain knowledge in anomaly detection include \e{contextual anomaly detection}, where anomality is conditioned on one or more contextual attributes \cite{song07conditional,li23explainable}, and \e{fair anomaly detection}, where the goal is to ensure that anomalies follow the same distribution over one or more sensitive attributes as normal records \cite{p20fair,zhang21towards,shekhar21fairod}.

\section{Proposal: monotonic anomaly detection}
\label{sec_dad}

As discussed in the introduction, we want to find a way to take into account \e{monotonic} attributes, for which we should only interpret extreme values in one direction as anomalous. Without loss of generality, we will assume that high values correspond to anomality --- any attributes for which the opposite applies can be transformed with a sign change.

\subsection{Modified Minkowski $p$-distance}

The anomaly detection algorithms CD, NND, LOF and ALP are all based on a choice of distance measure $d$ between various records. For tabular data, a typical choice for the distance $d$ is a form of Minkowski $p$-distance:

\begin{equation}
\label{eq_minkowski}
 d(y, x) = \size{y - x}_p := \Big(\sum_{1 \leq i \leq m} \lvert y_i - x_i \rvert^p\Big)^{\frac{1}{p}},
\end{equation}
for some $p \in (0, \infty)$.\footnote{In addition, Minkowski $\infty$-distance, also known as Chebyshev distance, is the limit case $p \ra \infty$, which resolves to $d(y, x) = \max_{i \leq m} \left\lvert y_i - x_i \right\rvert$.} In particular, the two most frequently used choices are 1-distance and 2-distance.\footnote{1-distance is also known as Boscovich, city-block or Manhattan distance. 2-distance is also known as Euclidean distance.}

Minkowski $p$-distance, applied to a test record $y$ and a training record $x$, is a function of the per-attribute differences $y_i - x_i$. When $y_i$ is larger than $x_i$, this contributes positive evidence to the anomality of $y$. And because the Minkowski $p$-distance takes the absolute value of the per-attribute differences, values of $y_i$ that are smaller than $x_i$ likewise contribute positive evidence for the anomality of $y$.

However, when we are dealing with a monotonic attribute, small values of $y_i$ are not an indication of anomality. Therefore, we propose instead to apply the ramp function\footnote{This is also known as the \e{rectifier} function, or \e{rectified linear unit} (ReLU), and is a popular activation function in neural networks \cite{fukushima69visual}.} $\max(0, y_i - x_i)$, which is asymmetric. This allows us to not interpret small values of $y_i$ as providing evidence for anomality.

In general, a dataset may contain a mixture of monotonic and ordinary attributes. Therefore, we propose the following, generalised form of Minkowski $p$-distance:

\begin{equation}
\label{eq_minkowski_generalised}
 d(y, x) = \brackets[\Big]{\sum_{1 \leq i \leq \mu} \max(0, {y_i - x_i)}^p + \sum_{\mu \leq i \leq m} \size{y_i - x_i}^p}^{\frac{1}{p}},
\end{equation}
where the attributes are ordered such that $x_1, x_2, \dots, x_{\mu}$ are the monotonic attribute values, and $x_{\mu + 1}, x_{\mu + 2}, \dots, x_{m}$ the non-monotonic attribute values.

Because this distance function is asymmetric, it is not a proper metric. Nonetheless, we can simply substitute it in all distance calculations of CD, NND, LOF and ALP, including between training records.

\subsection{SVM}

The anomaly score of SVM for a test record $y$ corresponds inversely to the following weighted sum:

\begin{equation}
\label{eq_svm}
 \sum_{x \in X} \alpha_x \kappa(y, x),
\end{equation}
where $X$ is the training set, $\kappa$ is a choice of kernel function measuring similarity, and the $\alpha_x$ are weights selected by the SVM algorithm. These weights can be equal to 0, and thus SVM effectively operates as a prototype selection algorithm.

The most common choice for $\kappa$ is the Gaussian kernel, which is the case $p = 2$ of the following class of kernels:

\begin{equation}
\label{eq_kernel}
 \kappa(y, x) = e^{-\size{y - x}_p^p/c},
\end{equation}
where the so-called kernel width $c$ is a hyperparameter that has to be chosen by the user. This class of kernels is also based on Minkowski $p$-distance, and we propose to substitute the modified Minkowski $p$-distance \eqref{eq_minkowski_generalised} into \eqref{eq_kernel} when calculating the anomaly score in \eqref{eq_svm}.\footnote{However, we will still use ordinary Minkowski $p$-distance in the original calculation of the weights $\alpha_i$, i.e., for the prototype selection.}

\subsection{IF}

IF is a tree ensemble algorithm, which fits $t$ random trees on random samples of size $\psi$ from the training data, for a choice of $t$ and $\psi$. Its anomaly score for a test record $y$ is:

\begin{equation}
\label{eq_if}
 2^{-\frac{1}{t}\sum_{i \leq t}h_i(y) / c(\psi)},
\end{equation}
where $h_i(y)$ is the path length of $y$ in tree $i$ and $c(\psi)$ is the expected path length in a random binary tree of $\psi$ records. The motivation for this definition is that an anomaly should on average require fewer splits than a normal record before it is isolated.

However, for monotonic attributes, a rightward split (corresponding to a higher attribute value) should not decrease the anomaly score. Therefore, our proposal is to calculate a modified path length, for which rightward splits on monotonic attributes are not counted, whereas leftward splits on monotonic attributes are counted double (such that the expected path length stays the same).

\section{Experimental setup}
\label{sec_experimental_setup}

We will now conduct a series of experiments on synthetic and real-life datasets to evaluate whether the approach proposed in this paper leads to better anomaly detection performance on datasets with monotonic attributes.

\subsection{Evaluation}

For the distance-based anomaly detectors, we compare our modified form of Minkowski $p$-distance \eqref{eq_minkowski_generalised} against ordinary Minkowski $p$-distance \eqref{eq_minkowski}. For IF, we compare our modified path lengths against normal path lengths. We evaluate anomaly detection performance using the area under the receiver operating characteristic (AUROC). This expresses the ability of an anomaly detector to separate anomalies from normal data, with 0.5 expressing no separation and 1 expressing complete separation. For the real-life datasets, we also conduct one-sided Wilcoxon signed-rank tests to evaluate, for each anomaly detector, the strength of the evidence that our proposal leads to increased performance.

\subsection{Synthetic datasets}

The synthetic datasets contain 1000 normal training records and 200 test records --- 100 normal and 100 anomalous. Each record consists of 10 attributes with randomly generated values. We consider two types of attributes:

\begin{itemize}
 \item Continuous attributes, where both normal and anomalous values follow a Gaussian distribution with standard deviation 1, centred on 0 and $a$ respectively. We let $a$ vary between 0 and 1 in steps of 0.1.
 \item Binary attributes, where both normal and anomalous values follow Bernoulli distributions with $p = 0.5 - 0.5 \cdot b$ and $p = 0.5 + 0.5 \cdot b$ respectively, for which we let $b$ vary between 0 and 0.5 in steps of 0.05.
\end{itemize}
For each value of $a$ and each value of $b$ we generate 100 datasets, and calculate the mean AUROC across these datasets.

\subsection{Real-life datasets}

Based on the accompanying documentation (and before starting experimentation), we have selected 12 real-life datasets from the UCI repository \cite{dua19uci} that can be approached as monotonic anomaly detection problems. The main properties of these datasets are described in the Appendix. Two of the datasets (\e{ai4i2020} and \e{post-operative}) contain a mixture of monotonic and ordinary attributes, while the other datasets only have monotonic attributes. Where necessary, we invert monotonic attributes to ensure that higher (rather than lower) values are indicative of anomality.

On these datasets, we perform 5-fold cross-validation on the normal records, creating at each iteration a test set by combining one fifth of the normal records with all anomalous records.

\subsection{Implementation details}

The anomaly detection algorithms in this paper have a number of hyperparameters which can in principle be tuned when there are anomalies available for validation purposes \cite{lenz22optimised}. In general, however, this is not the case in the semi-supervised setting, so we will use sensible default values, as established in \cite{lenz21average}. For NND, we use the weighted variant introduced in \cite{lenz23fuzzy}. For SVM and IF, we rely on the implementations provided by the scikit-learn library \cite{pedregosa11scikitlearn}.

Because the distance-based algorithms depend on the relative scale of the attributes, we normalise the real-life datasets by dividing each attribute by the semi-interquartile range of the normal training values. Correspondingly, we let CD be the distance to the midhinge.

A link to the code for generating the results is found at the end of this paper.

\begin{figure}
\centering
\includegraphics[width=.9\linewidth]{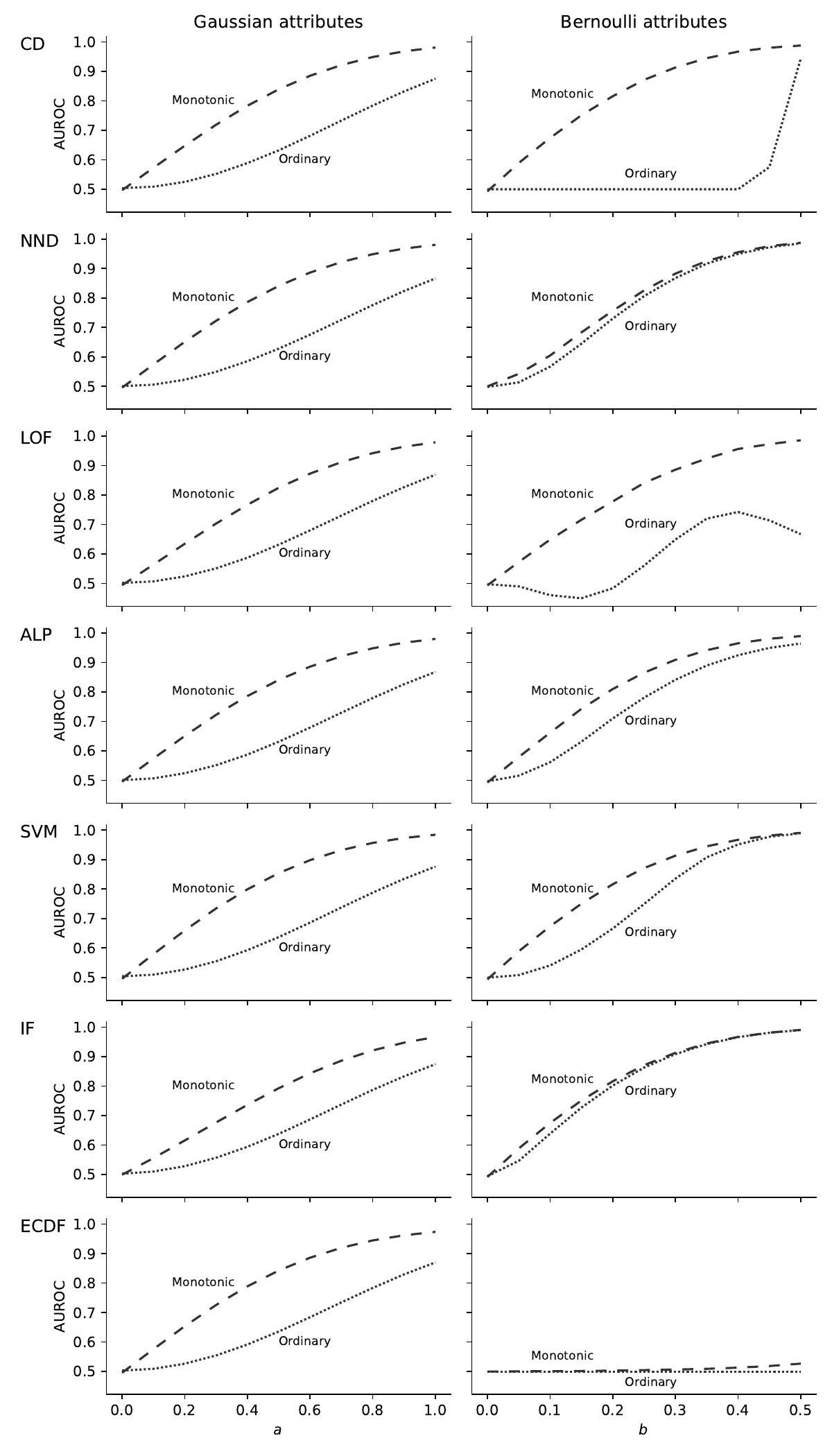}
\caption{Mean AUROC obtained by the monotonic and ordinary variants of each anomaly detector (using 1-distance) on synthetic datasets with Gaussian and Bernoulli attributes. $a$: distance between the distributions of anomalous and normal attribute values; $b$: difference between the probability of a positive anomalous and a positive normal attribute value.}
\label{fig_synthetic_results_1}
\end{figure}

\begin{table}
\centering
\caption{Mean cross-validation AUROC for the ordinary (O) and monotonic (M) variants of each anomaly detector. Distance-based anomaly detectors use 1-distance (Boscovich). \textbf{{Bold}}: higher value (before rounding).}
\label{tab_organic_auroc_1}
\begin{tabular}{lp{.065\linewidth}p{.085\linewidth}p{.065\linewidth}p{.085\linewidth}p{.065\linewidth}p{.085\linewidth}p{.065\linewidth}p{.065\linewidth}}
\toprule
Dataset & \multicolumn{2}{l}{CD} & \multicolumn{2}{l}{NND} & \multicolumn{2}{l}{LOF} & \multicolumn{2}{l}{ALP} \\
 & O & M & O & M & O & M & O & M \\
\midrule
ai4i2020 & 0.800 & \bftab 0.861 & 0.823 & \bftab 0.922 & 0.867 & \bftab 0.909 & 0.877 & \bftab 0.924 \\
diabetes-risk & \bftab 0.910 & 0.798 & \bftab 0.971 & 0.923 & 0.745 & \bftab 0.920 & 0.898 & \bftab 0.928 \\
fertility & 0.493 & \bftab 0.541 & 0.602 & \bftab 0.653 & 0.534 & \bftab 0.605 & 0.582 & \bftab 0.636 \\
heart-failure & 0.740 & \bftab 0.774 & 0.715 & \bftab 0.769 & \bftab 0.736 & 0.715 & 0.734 & \bftab 0.766 \\
phishing-websites & 0.645 & \bftab 0.752 & 0.901 & \bftab 0.927 & 0.851 & \bftab 0.902 & 0.927 & \bftab 0.935 \\
post-operative & 0.473 & \bftab 0.512 & 0.476 & \bftab 0.504 & 0.522 & \bftab 0.527 & 0.460 & \bftab 0.485 \\
qualitative-bankruptcy & \bftab 0.999 & 0.999 & \bftab 1.000 & 1.000 & \bftab 1.000 & 1.000 & \bftab 1.000 & 1.000 \\
south-german-credit & 0.632 & \bftab 0.717 & 0.648 & \bftab 0.718 & 0.638 & \bftab 0.675 & 0.648 & \bftab 0.714 \\
thoraric-surgery & 0.569 & \bftab 0.598 & 0.597 & \bftab 0.624 & \bftab 0.607 & 0.542 & \bftab 0.634 & 0.622 \\
wdbc & 0.953 & \bftab 0.967 & 0.950 & \bftab 0.976 & 0.950 & \bftab 0.954 & 0.957 & \bftab 0.981 \\
wisconsin & \bftab 0.995 & 0.995 & \bftab 0.995 & 0.994 & 0.692 & \bftab 0.990 & 0.870 & \bftab 0.995 \\
wpbc & 0.557 & \bftab 0.627 & 0.570 & \bftab 0.625 & 0.555 & \bftab 0.582 & 0.537 & \bftab 0.654 \\[.25\baselineskip]
\e{Mean} & 0.731 & \bftab 0.762 & 0.771 & \bftab 0.803 & 0.725 & \bftab 0.777 & 0.760 & \bftab 0.803 \\
\midrule
Dataset & \multicolumn{2}{l}{SVM} & \multicolumn{2}{l}{IF} & \multicolumn{2}{l}{ECDF} \\
 & O & M & O & M & O & M \\
\midrule
ai4i2020 & 0.862 & \bftab 0.863 & 0.801 & \bftab 0.874 & 0.846 & \bftab 0.905 \\
diabetes-risk & \bftab 0.974 & 0.840 & \bftab 0.930 & 0.874 & \bftab 0.920 & 0.893 \\
fertility & 0.484 & \bftab 0.553 & 0.606 & \bftab 0.644 & 0.460 & \bftab 0.577 \\
heart-failure & 0.734 & \bftab 0.771 & 0.716 & \bftab 0.725 & 0.720 & \bftab 0.780 \\
phishing-websites & \bftab 0.868 & 0.840 & 0.668 & \bftab 0.729 & 0.750 & \bftab 0.810 \\
post-operative & 0.452 & \bftab 0.486 & \bftab 0.471 & 0.468 & 0.473 & \bftab 0.497 \\
qualitative-bankruptcy & \bftab 1.000 & 1.000 & 0.979 & \bftab 0.983 & \bftab 1.000 & \bftab 1.000 \\
south-german-credit & 0.637 & \bftab 0.711 & 0.602 & \bftab 0.674 & 0.618 & \bftab 0.690 \\
thoraric-surgery & 0.573 & \bftab 0.608 & 0.597 & \bftab 0.634 & 0.574 & \bftab 0.616 \\
wdbc & 0.954 & \bftab 0.977 & 0.960 & \bftab 0.972 & 0.929 & \bftab 0.963 \\
wisconsin & 0.995 & \bftab 0.995 & 0.995 & \bftab 0.995 & \bftab 0.990 & 0.990 \\
wpbc & 0.565 & \bftab 0.631 & 0.533 & \bftab 0.599 & 0.487 & \bftab 0.602 \\[.25\baselineskip]
\e{Mean} & 0.758 & \bftab 0.773 & 0.738 & \bftab 0.764 & 0.731 & \bftab 0.777 \\
\bottomrule
\end{tabular}
\end{table}

\begin{table}
\centering
\caption{$p$-values from a one-sided Wilcoxon signed-rank test of the monotonic vs. the ordinary variants.}
\label{tab_organic_p_values}
\begin{tabular}{p{.300\linewidth}p{.08\linewidth}p{.08\linewidth}p{.08\linewidth}p{.08\linewidth}p{.08\linewidth}p{.08\linewidth}p{.08\linewidth}}
\toprule
 & CD & NND & LOF & ALP & SVM & IF & ECDF \\
\midrule
1-distance (Boscovich) & 0.030 & 0.011 & 0.025 & 0.003 & 0.058 & 0.011 & 0.005 \\
2-distance (Euclidean) & 0.017 & 0.011 & 0.120 & 0.002 & 0.050 &  &  \\
\bottomrule
\end{tabular}
\end{table}

\section{Results and discussion}
\label{sec_results}

For reasons of space, we focus on the results with 1-distance, since the results with 2-distance are very similar. The performance of the anomaly detectors on synthetic and real-life datasets is displayed in, respectively, Figure~\ref{fig_synthetic_results_1} and Table~\ref{tab_organic_auroc_1}, and the $p$-values of the significance tests for the real-life datasets in Table~\ref{tab_organic_p_values}.

We first note that none of the methods performed better than chance on the \e{post-operative} dataset, making this essentially unsuitable as an anomaly detection problem. However, we have chosen not to exclude it in order not to bias our dataset selection.

A second observation is that the ordinary variants of CD, LOF and ECDF are ill-equipped for dealing with binary attributes (Figure~\ref{fig_synthetic_results_1}, right column), and that for CD and LOF, this is alleviated by the monotonic variants that we have proposed.

More generally, we find that the monotonic variants of the anomaly detection algorithms outperform the ordinary variants on our selection of datasets with monotonic attributes. On the real-life datasets, the difference is significant ($p < 0.05$) except for SVM with 1-distance and LOF with 2-distance. On those real-life datasets where the monotonic variants fail to outperform the ordinary variants, the difference is very small and/or the AUROC scores are very low for both variants.

There is one notable exception: the \e{diabetes-risk} dataset. We believe that this is due to the fact that the attributes of this dataset encode diabetes symptoms, and that the normal records do not represent average healthy people, but non-diabetes patients who nevertheless display symptoms of diabetes. This undermines the monotonicity of this dataset --- diabetes patients may not necessarily have more symptoms than this particular group of other patients, but rather \e{different combinations} of symptoms. Indeed, when we look at the mean attribute values of normal and anomalous records in this dataset, we find that for five of the fourteen attributes, the anomalous records have lower or only slightly higher mean values than the normal records.

Lastly, we note that the proposed monotonic variants of the two best distance-based algorithms (NND and ALP) outperform the only monotonic proposal from the literature (ECDF), with, respectively, $p = 0.29$ and $p = 0.34$ from a one-sided Wilcoxon signed-rank test, corrected with the Benjamini-Hochberg method \cite{benjamini95controlling} to control the false discovery rate.

\section{Conclusion}
\label{sec_conclusion}

In this paper, we have introduced monotonic anomaly detection, a new problem setting that incorporates the domain knowledge that only high (or only low) values of certain attributes are indicative of anomality. Distance-based anomaly detectors can take this monotonicity into account by using an asymmetric distance function. For this purpose, we have proposed a variant of the commonly used family of Minkowski $p$-distances that incorporates the ramp function. For the non-distance-based Isolation Forest algorithm, we have similarly proposed a modification to the calculation of path lengths.

In a series of experiments on synthetic and real-life datasets, we have shown that these monotonic variants generally increase anomaly detection performance in datasets with monotonic attributes. On the \e{diabetes-risk} dataset, the monotonic variant of some algorithms achieved lower performance, but we have argued that this is due to the fact that on closer inspection, not all of the attributes in this dataset are really monotonic.

We believe that the most important contribution of our proposal is that it allows practitioners to use an anomaly detection model that aligns more closely with their domain knowledge than a general-purpose algorithm. This has three advantages. Firstly, as we have demonstrated, it will generally increase detection performance, reducing the workload of the practitioner. Secondly, as with the \e{diabetes-risk} dataset, when it does not increase performance, it may prompt the practitioner to question their understanding of the problem at hand, potentially leading to new insights. And thirdly, a model that more closely aligns with the practitioner's domain knowledge is arguably more trustworthy even in the event that it should lead to slightly lower detection performance.

\begin{credits}

\subsubsection{Data and code} Datasets and the code to reproduce our experiments are available at \url{https://liacs.leidenuniv.nl/~lenzou/code/lenz-2026-monotonic.tar.gz}.

\subsubsection{\ackname} This publication is part of the project Digital Twin with project number P18-03 of the research programme TTW Perspective, which is (partly) financed by the Dutch Research Council (NWO).

\end{credits}

\appendix

\section*{Appendix 1: Real-life datasets}
Table~\ref{tab_statistics} lists the main properties of the real-life datasets in our experiment. The attributes of these datasets express machine operating conditions (\e{ai4i2020}), medical symptoms, comorbidities and lifestyle factors (\e{diabetes-risk}, \e{fertility}, \e{heart-failure}, \e{post-operative} and \e{thoraric-surgery}), tumor characteristics (\e{wdbc}, \e{wisconsin} and \e{wpbc}), and risk indicators provided by experts (\e{phishing-websites}, \e{qualitative-bankruptcy} and \e{south-german-credit}). For \e{ai4i2020} and \e{post-operative}, respectively four out of six and five out of eight attributes are monotonic, while the other datasets only have monotonic attributes.

\begin{table}
\caption{Real-life datasets used in the experiment. $m$: number of attributes.}
\label{tab_statistics}
{\begin{tabular}{llrp{.2\linewidth}rp{.21\linewidth}r}
\toprule
\multicolumn{1}{l}{Dataset} & Source & \multicolumn{2}{l}{Records} & \multicolumn{2}{l}{Anomalies} & $m$ \\
\midrule
ai4i2020 & \cite{matzka20explainable} & 10\,000 & Simulated machine operation & 339 & Five different failure modes & 6 \\
diabetes-risk & \cite{islam20likelihood} & 520 & Patients with diabetes symptoms & 320 & Actual diabetes patients & 14 \\
fertility & \cite{gil12predicting} & 100 & Sperm samples & 12 & `Altered' samples & 8 \\
heart-failure & \cite{ahmad17survival} & 299 & Heart patients & 96 & Deaths & 11 \\
phishing-websites & \cite{thabtah16dynamic} & 11\,055 & Websites & 4898 & Phishing websites & 30 \\
post-operative & \cite{budihardjo91program} & 87 & Surgery patients & 63 & Patients who had to stay in hospital & 8 \\
qualitative-bankruptcy & \cite{kim03discovery} & 250 & Companies & 107 & Bankruptcies & 6 \\
south-german-credit & \cite{gromping19south} & 1000 & Bank credits & 300 & Bad credits & 20 \\
thoraric-surgery & \cite{zieba14boosted} & 470 & Lung cancer surgery patients & 70 & Deaths & 16 \\
wdbc & \cite{street92nuclear} & 569 & Breast tumor samples & 212 & Malignancy & 30 \\
wisconsin & \cite{wolberg90multisurface} & 683 & Fine-needle aspirates of breast tumors & 239 & Malignancy & 9 \\
wpbc & \cite{street96individual} & 138 & Breast cancer surgery patients & 28 & Recurrence & 32 \\
\bottomrule
\end{tabular}}
\end{table}

\section*{Appendix 2: Results with 2-distance (Euclidean)}

For the distance-based anomaly detectors, Fig.~\ref{fig_synthetic_results_2} and Table~\ref{tab_organic_auroc_2} display the results obtained with 2-distance (Euclidean). These are very similar to the results with 1-distance (Boscovich), displayed in  Fig.~\ref{fig_synthetic_results_1} and Table~\ref{tab_organic_auroc_1}.

\begin{figure}
\centering
\includegraphics[width=.9\linewidth]{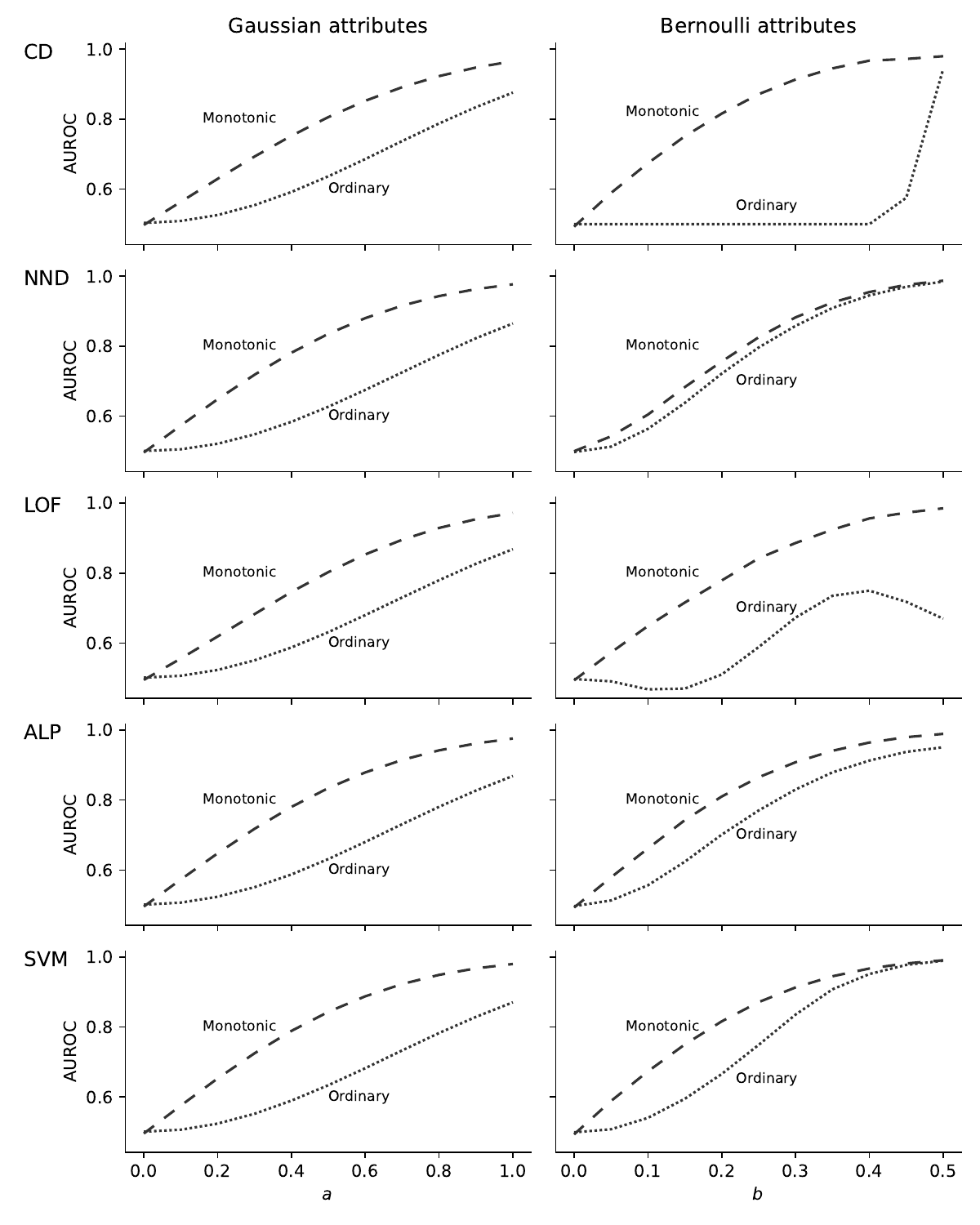}
\caption{Mean AUROC obtained by the monotonic and ordinary variants of each anomaly detector (using 2-distance) on synthetic datasets with Gaussian and Bernoulli attributes. $a$: distance between the distributions of anomalous and normal attribute values; $b$: difference between the probability of a positive anomalous and a positive normal attribute value.}
\label{fig_synthetic_results_2}
\end{figure}

\begin{table}
\centering
\caption{Mean cross-validation AUROC for the ordinary (O) and monotonic (M) variants of each distance-based anomaly detector with 2-distance (Euclidean). \textbf{{Bold}}: higher value (before rounding).}
\label{tab_organic_auroc_2}
\begin{tabular}{lp{.060\linewidth}p{.075\linewidth}p{.060\linewidth}p{.075\linewidth}p{.060\linewidth}p{.075\linewidth}p{.060\linewidth}p{.075\linewidth}p{.060\linewidth}p{.060\linewidth}}
\toprule
Dataset & \multicolumn{2}{l}{CD} & \multicolumn{2}{l}{NND} & \multicolumn{2}{l}{LOF} & \multicolumn{2}{l}{ALP} & \multicolumn{2}{l}{SVM} \\
 & O & M & O & M & O & M & O & M & O & M \\
\midrule
ai4i2020 & 0.798 & \bftab 0.855 & 0.821 & \bftab 0.919 & 0.862 & \bftab 0.904 & 0.873 & \bftab 0.919 & 0.859 & \bftab 0.874 \\
diabetes-risk & \bftab 0.910 & 0.798 & \bftab 0.973 & 0.928 & 0.733 & \bftab 0.923 & 0.846 & \bftab 0.933 & \bftab 0.974 & 0.818 \\
fertility & 0.523 & \bftab 0.537 & 0.554 & \bftab 0.647 & \bftab 0.508 & 0.500 & 0.511 & \bftab 0.632 & 0.496 & \bftab 0.559 \\
heart-failure & 0.732 & \bftab 0.771 & 0.720 & \bftab 0.771 & \bftab 0.737 & 0.689 & 0.732 & \bftab 0.768 & 0.710 & \bftab 0.779 \\
phishing-websites & 0.672 & \bftab 0.753 & 0.911 & \bftab 0.928 & 0.833 & \bftab 0.907 & 0.920 & \bftab 0.938 & \bftab 0.877 & 0.857 \\
post-operative & 0.486 & \bftab 0.522 & 0.499 & \bftab 0.511 & \bftab 0.531 & 0.521 & 0.454 & \bftab 0.490 & 0.463 & \bftab 0.476 \\
qualitative-bankruptcy & 0.999 & \bftab 1.000 & 1.000 & \bftab 1.000 & 1.000 & \bftab 1.000 & 1.000 & \bftab 1.000 & \bftab 1.000 & 1.000 \\
south-german-credit & 0.631 & \bftab 0.710 & 0.644 & \bftab 0.722 & 0.640 & \bftab 0.684 & 0.646 & \bftab 0.716 & 0.643 & \bftab 0.723 \\
thoraric-surgery & 0.541 & \bftab 0.592 & 0.592 & \bftab 0.624 & \bftab 0.592 & 0.535 & \bftab 0.622 & 0.620 & 0.575 & \bftab 0.606 \\
wdbc & 0.953 & \bftab 0.961 & 0.947 & \bftab 0.970 & 0.948 & \bftab 0.951 & 0.953 & \bftab 0.979 & 0.948 & \bftab 0.975 \\
wisconsin & 0.995 & \bftab 0.995 & \bftab 0.994 & 0.994 & 0.764 & \bftab 0.986 & 0.920 & \bftab 0.992 & 0.992 & \bftab 0.993 \\
wpbc & 0.589 & \bftab 0.627 & 0.581 & \bftab 0.625 & 0.567 & \bftab 0.604 & 0.532 & \bftab 0.647 & 0.574 & \bftab 0.616 \\[.25\baselineskip]
\e{Mean} & 0.736 & \bftab 0.760 & 0.770 & \bftab 0.803 & 0.726 & \bftab 0.767 & 0.751 & \bftab 0.803 & 0.759 & \bftab 0.773 \\
\bottomrule
\end{tabular}
\end{table}

\newpage

\bibliographystyle{splncs04}
\bibliography{20250527_mad}

\end{document}